\documentclass[journal]{IEEEtran}

\usepackage{graphicx}
\usepackage{amsmath}
\usepackage{cite}
\usepackage{hyperref}
\usepackage{caption}
\usepackage{subcaption}
\usepackage{color}
\usepackage{amsfonts}
\usepackage{bbm}
\usepackage{algorithm}
\usepackage{algpseudocode}
\usepackage{mathtools}
\usepackage{tabularx}
\usepackage{amsthm}

\title{An Automated Tip-and-Cue Framework for Optimized Satellite Tasking and Visual Intelligence}

\author{Gil Weissman, Amir Ivry,~\IEEEmembership{Senior Member,~IEEE,} Israel Cohen,~\IEEEmembership{Fellow,~IEEE}
\thanks{This research was supported by the Israel Science Foundation (grant no. 1449/23) and the Pazy Research Foundation.}
\thanks{The authors are with the Andrew and Erna Viterbi Faculty of Electrical and Computer Engineering, Technion-Israel Institute of Technology, Haifa 3200003, Israel (e-mail: gil.weissman@campus.technion.ac.il; sivry@technion.ac.il; icohen@ee.technion.ac.il).}
}

\begin{document}

\maketitle

\begin{abstract}
The proliferation of satellite constellations, coupled with reduced tasking latency and diverse sensor capabilities, has expanded the opportunities for automated Earth observation. This paper introduces a fully automated Tip-and-Cue framework designed for satellite imaging tasking and scheduling. In this context, \textit{tips} are generated from external data sources or analyses of prior satellite imagery, identifying spatiotemporal targets and prioritizing them for downstream planning. Corresponding \textit{cues} are the imaging tasks formulated in response, which incorporate sensor constraints, timing requirements, and utility functions. The system autonomously generates candidate tasks, optimizes their scheduling across multiple satellites using continuous utility functions that reflect the expected value of each observation, and processes the resulting imagery using artificial-intelligence-based models, including object detectors and vision-language models. Structured visual reports are generated to support both interpretability and the identification of new insights for downstream tasking. The efficacy of the framework is demonstrated through a maritime vessel tracking scenario, utilizing AIS (Automatic Identification System) data for trajectory prediction, targeted observations, and the generation of actionable outputs. Maritime vessel tracking is a widely researched application, often used to benchmark novel approaches to satellite tasking, forecasting, and analysis. The system is extensible to broader applications such as smart-city monitoring and disaster response, where timely tasking and automated analysis are critical.
\end{abstract}


\section{Introduction}
\label{sec:introduction}

The increasing availability of satellite constellations, combined with reduced tasking latency and growing sensor diversity, has significantly enhanced the potential for responsive Earth observation. This evolution enables the development of automated frameworks that can dynamically generate, schedule, and analyze satellite imaging tasks. However, managing task allocation across heterogeneous satellite assets while maintaining high spatial and temporal relevance remains a complex optimization challenge.

Despite substantial advances in anomaly detection, satellite tasking, and post-acquisition analysis, most existing efforts address these phases in isolation. For example, anomaly detection methods from AIS data focus on identifying irregular behaviors, while scheduling methods address efficient resource allocation under constraints. Post-acquisition analysis, on the other hand, emphasizes object detection or semantic interpretation. Yet, a significant gap remains in unifying these phases under a single, rigorous framework. Current approaches rarely support an end-to-end loop where anomaly-based triggers directly inform cue generation, which in turn feeds continuous-time scheduling, and where post-acquisition analysis can automatically refine or generate new tips. This lack of integration limits responsiveness, scalability, and the ability to adaptively plan observations under evolving priorities.

Prior work on anomaly detection from AIS data ranges from probabilistic models~\cite{pallotta2013vessel} to deep learning approaches such as trajectory modeling~\cite{nguyen2022geotracknet}, AIS dropout classification~\cite{singh2020mlais}, and transformer-based signal irregularity detection~\cite{bernabe2024shutdown}. Cross-modal systems that fuse AIS with imagery have shown success in detecting dark vessels or resolving mismatches~\cite{wang2024fusion, chen2025sviadf}. Several efforts have used anomaly-based triggers for cues~\cite{harun2022multisensor}, but most of these are limited to detection and do not propagate insights downstream. Our framework extends state-of-the-art models, such as TrAISformer~\cite{traisformer2024}, with predictive scoring, spatial-temporal filtering, and cue generation, while explicitly supporting integration into real-time scheduling pipelines.

For satellite tasking, traditional approaches are based on analytical methods~\cite{lemaitre2002agile, globus2003scheduling}, heuristics~\cite{zhang2014multi, wu2022data}, and reinforcement learning~\cite{huang2021revising, ou2023deep, mercado2025energy}. Hybrid methods~\cite{wu2022ensemble, han2022simulated} and recent graph-based or quantum-based schedulers~\cite{zhang2024grlr, marchioli2025quantum} improve scalability and adaptivity. However, many of these assume discrete time slots, fixed task windows, and lack closed-loop or differentiable utility formulations. Our work proposes a novel projected gradient-based scheduler for continuous acquisition times, which incorporates soft constraints, differentiable utility maximization, and real-time cue prioritization across multiple satellites and heterogeneous sensors.

Post-acquisition analysis in most frameworks is limited to object detection or land classification~\cite{zhu2017deep, gao2020multiscale}, with minimal feedback to tasking. Recent work in the semantic interpretation of EO (Electro-Optic) imagery includes systems such as RemoteCLIP~\cite{liu2024remoteclip}, MapGlue~\cite{wu2025mapglue}, and image anomaly detection~\cite{wang2021unified}. Meanwhile, tools like EOAD~\cite{castillo2021earth} detect field-level changes. However, few leverage vision-language models or produce human-readable alerts. Feedback from image-based anomalies in re-tasking remains rare. In contrast, our framework integrates AI-based enrichment and VLM-generated reports into the decision loop, enabling refined tips and iterative cue generation from imagery.

This paper addresses the above gap by presenting a fully automated Tip-and-Cue framework for satellite imaging tasking and scheduling. In this paradigm, a \textit{tip} refers to an alert or recommendation generated from external sources, such as AIS feeds, or from the analysis of previously acquired satellite images and metadata. Each tip identifies a spatiotemporal region of interest, a discrete detection time, and a priority score for downstream tasking. A \textit{cue} is defined as the imaging task formulated in response to a tip, incorporating sensor constraints, timing feasibility, and a utility function defined over time to evaluate the acquisition value. This closed-loop mechanism facilitates adaptive observation planning based on evolving inputs and previous outcomes.

The proposed framework introduces: predictive tip scoring from heterogeneous data streams, continuous-time utility-driven scheduling with soft constraints and projected gradient optimization, and post-acquisition semantic enrichment with automated feedback into tip generation. We demonstrate the efficiency of this framework using a combination of dynamic AIS-derived tips and static areas of interest over the U.S. East Coast, employing various EO satellites, such as SKYSAT-C11, SKYSAT-C15, and JILIN-1-GF03D50, to optimize acquisition planning under realistic constraints. The SkySat constellation provides submeter imagery with high revisit capability~\cite{saunier2022skysat}, while JILIN‑1’s GF03D series supports high-resolution Earth observation in China’s commercial constellation~\cite{yu2020board, jianing2022research}.

\section{Framework Overview}
\label{sec:framework_overview}
\begin{figure}[t]
  \centering
  \includegraphics[width=0.75\linewidth]{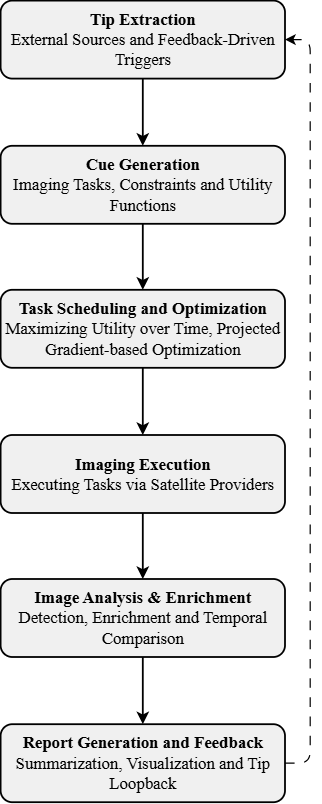}
  \caption{Architecture of the Tip-and-Cue Satellite Imaging Framework.}
  \label{fig:pipeline}
\end{figure}

The proposed pipeline is a fully automated framework for satellite imaging tasking and scheduling, following the Tip-and-Cue paradigm. It enables adaptive, end-to-end coordination across heterogeneous satellite constellations by continuously generating, prioritizing, executing, and analyzing imaging tasks. A diagram of this pipeline is shown in Fig.~\ref{fig:pipeline}.

The system consists of the following six stages:

\begin{enumerate}
\item \textbf{Tip Extraction:} Tips are derived from two sources: external streams (e.g., AIS, weather alerts) and feedback-based triggers generated from prior satellite imagery and system analysis. These include anomaly detection, semantic drift, and contextual relevance, which are identified during report generation. Each tip identifies a spatiotemporal region of interest, a discrete detection time, and is assigned a base priority score to guide the generation of downstream tasks.
    
\item \textbf{Cue Generation:} Each tip is converted into one or more candidate imaging tasks (cues), formally defined by geographic location, temporal window, and explicit imaging constraints, such as the required resolution, incidence angle, or sensor modality. Each cue is coupled with a continuous utility function that quantifies the expected value of acquiring the observation over time.

\item \textbf{Task Scheduling and Optimization:} The system performs a constrained optimization over all continuous utility functions defined by the cues, taking into account the limited imaging capacity of the available satellite platforms. The objective is to select the most valuable subset of cues and assign them to optimal and feasible acquisition times, subject to sensor constraints, visibility geometry, and minimum temporal separation. A projected gradient descent method is applied to identify local optimal schedules under these feasibility constraints efficiently.

\item \textbf{Imaging Execution:} Optimized imaging schedules are submitted to satellite operators. The system supports various constellations with different orbital characteristics, imaging modes, and tasking interfaces.

\item \textbf{Image Analysis and Enrichment:} Acquired imagery is processed using AI models, including object detectors and vision-language models (VLMs). Structured outputs, such as bounding boxes, captions, and semantic tags, are compared with historical imagery and external data sets to validate observations and extract higher-level insights.

\item \textbf{Report Generation and Feedback:} Results are compiled into visual reports with detections, context summaries, and comparative analyses. These reports support both human interpretation and automated reasoning, feeding back into the tip extraction stage for continuous, adaptive operation. This feedback complements externally sourced tips and enables re-tasking based on recent observations.
\end{enumerate}

\section{Methodology}
\label{sec:methodology}

The proposed methodology follows a modular pipeline consisting of six core components as described in Section~\ref{sec:framework_overview}. Each component is designed to be general-purpose and fully automated, enabling scalable deployment across different satellite constellations and sensing objectives.

\subsection{Tip Extraction}
\label{subsec:tip_extraction}

The tip extraction module identifies spatiotemporal events of interest from external or internal data sources. Each tip \( \tau_i \) is defined as:
\begin{equation}
\tau_i = \bigl(\mathcal{P}_i, n_i, s_i\bigr),
\end{equation}
where \( \mathcal{P}_i \subset \mathbb{R}^2 \) is a spatial anchor region, \( n_i \) is the discrete detection time, and \( s_i \in [0, 1] \) is a base priority score for downstream scheduling.

Tip generation is modeled as a binary decision over the current observation \(\boldsymbol{{d}_{n_i}}\) and historical context \(\boldsymbol{{H}_{n_i}}\):
\begin{equation}
f_{\tau_i}\bigl(\boldsymbol{d_{n_i}}, \boldsymbol{H_{n_i}}\bigr) =
\begin{cases}
1, & \text{if } \phi_{j,i} \coloneqq \phi_j\bigl(\boldsymbol{d_{n_i}}, \boldsymbol{H_{n_i}}\bigr) > \theta_j\\[6pt]
0, & \text{otherwise.}
\end{cases},
\end{equation}

where \( \phi_j(\cdot) \) is a modality-specific anomaly score, and \( \theta_j \) is its corresponding threshold. This framework unifies heterogeneous signal modalities under a common detection logic.

\vspace{0.2em}
\textit{External Sources:} Structured data streams such as AIS are monitored continuously. At each time \( n_i \), a model predicts the vessel's position \( \hat{\mathbf{x}}_i(n_i) \) from its recent trajectory \( \boldsymbol{H_{n_i}} \). A tip is triggered if the prediction error exceeds a threshold:
\begin{equation}
f_{\tau_i}\bigl(\boldsymbol{d_{n_i}}, \boldsymbol{H_{n_i}}\bigr) =
\begin{cases}
1, & \text{if } \phi_{\mathrm{AIS},i} > \theta_{\mathrm{AIS}}\\[6pt]
0, & \text{otherwise.}
\end{cases},
\end{equation}

Trajectory-based anomaly detection follows the paradigm introduced in~\cite{nguyen2022geotracknet}, where learned models flag deviations from expected movement patterns. Our formulation supports anticipatory anomaly detection by detecting forecast divergence rather than solely retrospective outliers~\cite{singh2020machine, ribeiro2023ais}.

\vspace{0.3em}
\textit{Feedback-Based Triggers:} Internally generated tips arise from automated analysis of previously acquired imagery. An autoencoder model produces an embedding vector \(\boldsymbol{z_{n_i}}\), which is compared to the mean historical embedding \(\boldsymbol{\bar{z}_{{H}_{n_i}}}\). A tip is raised if the cosine distance exceeds a semantic drift threshold:
\begin{equation}
\phi_{\mathrm{AE},i} = d_{\mathrm{cos}}\bigl(\boldsymbol{z_{n_i}}, \boldsymbol{\bar{z}_{{H}_{n_i}}}\bigr),
\qquad
f_{\tau_i} = \mathbbm{1}\bigl(\phi_{\mathrm{AE},i} > \theta_{\mathrm{AE}}\bigr),
\end{equation}
where \( \mathbbm{1}(\cdot) \) denotes the indicator function.
This enables integration of vision-language models and caption-based anomaly detection~\cite{wang2021unified}, expanding the range of observable events to semantic changes and content deviations. It also sets the foundation for closed-loop image-driven retasking.

\vspace{0.3em}
\textit{Tip Scoring:} Each tip receives a scalar score \( s_i \) defined by a weighted combination of deviation and urgency:
\begin{equation}
s_i = \alpha\, s_{\mathrm{dev},i} + (1 - \alpha)\, s_{\mathrm{urg},i},
\qquad \alpha \in [0, 1],
\end{equation}
where \( s_{\mathrm{dev},i} \) reflects deviation from expected behavior or content, and \( s_{\mathrm{urg},i} \) encodes time sensitivity.  The balance parameter \( \alpha \) is application-specific.

In the maritime vessel scenario, for example, let:
\begin{equation}
s_{\mathrm{dev},i} = 1 - \frac{\theta_{\mathrm{AIS}}}{\phi_{\mathrm{AIS},i}}.
\end{equation}
quantify how abnormal the prediction error is relative to the threshold.

\begin{figure}[t]
  \centering
  \includegraphics[width=\linewidth]{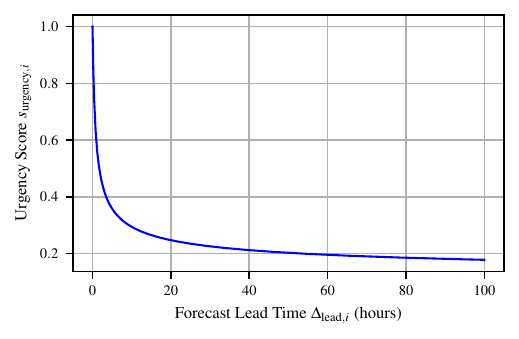}
  \caption{Urgency score as a function of the lead time \( \Delta_{\text{lead},i} \). The curve illustrates how urgency decreases logarithmically as forecast confidence time increases.}
  \label{fig:surgency_plot}
\end{figure}

Let \( \Delta_{\text{lead},i} \) denote the forecast lead time (in hours) required for this vessel type and location to ensure a 95\% confidence level in prediction. Then,
\begin{equation}
s_{\text{urg},i} = \frac{1}{1 + \log(1 + \Delta_{\text{lead},i})},
\end{equation}
quantifies the urgency of imaging this vessel in relation to the empirical performance of the prediction model. The functional behavior of \( s_{\text{urg},i} \) is shown in Figure~\ref{fig:surgency_plot}.

Since both \( s_{\text{dev},i} \in [0,1] \) and \( s_{\text{urg},i} \in [0,1] \), their convex combination \( s_i \in [0,1] \) as well.

The final tip score is computed as:
\begin{equation}
s_i = \alpha \left( 1 - \frac{\theta_{\text{AIS}}}{\phi_{\text{AIS},i}} \right) 
+ \frac{1 - \alpha}{1 + \log(1 + \Delta_{\text{lead},i})}, 
\quad \alpha \in [0,1]
\label{eq:scoring_AIS}.
\end{equation}
This formulation balances the intensity of deviation with temporal recency, enabling the robust prioritization of candidate tips for downstream scheduling.

\subsection{Cue Generation}
\label{subsec:cue_generation}

Given a tip \( \tau_i  \), the cue generation module constructs one or more candidate imaging tasks by assigning a continuous utility function over time and, when necessary, forecasting the spatial region of interest.

For dynamic targets (e.g., moving vessels), the spatial footprint is modeled as a time-dependent function \( \mathcal{P}_i(t) \), derived from predictive tracking or trajectory models. For static targets, this remains fixed as \( \mathcal{P}_i(t) = \mathcal{P}_i \).

This approach supports moving target observation scenarios as studied in~\cite{liu2024learning, cong2025autonomous}, where both the target state and observation utility evolve with time. In maritime contexts, AIS-driven forecasts are particularly well-suited for modeling \( \mathcal{P}_i(t) \), as demonstrated in~\cite{traisformer2024}.

Each candidate task is assigned a utility function \({u_i(t) \in [0, 1]}\), representing the relative value of acquiring an observation at time \( t \). This function is defined as:
\begin{equation}
u_i(t) = s_i \,\psi_i(t),
\end{equation}
where \( s_i \) is the base priority score derived during tip extraction, and \( \psi_i(t) \) is a differentiable and globally \( L_i \)-Lipschitz smooth function for some \( L_i \), capturing the time-dependent desirability. The function \( \psi_i(t) \) is designed to be independent of satellite observability and may incorporate factors such as urgency decay, expected weather conditions, or model confidence.

This decoupling aligns with recent frameworks that separate task utility modeling from platform constraints~\cite{jacquet2024earth}, enabling better modularity in scheduling pipelines.

For example, we consider a utility curve that rises to a peak and then decays symmetrically over time. This may represent scenarios where re-observation becomes increasingly important after the initial detection, up to an optimal point, after which prediction uncertainty or diminishing relevance reduces its utility.
We define:
\begin{equation}
\psi_i(t) = \exp\left[-\left(\frac{t - t_{\mathrm{peak}}}{\sigma_i}\right)^2\right],
\label{eq:gaussian_psi}
\end{equation}
where \( t_{\mathrm{peak}} \in \mathbb{R} \) denotes the time of maximum utility, and \( \sigma_i \in \mathbb{R}_+ \) controls the temporal spread around the peak. The utility reaches its maximum value \( s_i \) at \( t = t_{\mathrm{peak}} \) and decays smoothly before and after that point.

Alternatively, specific tasks, such as those triggered by prediction-based anomaly detection, may benefit from a utility that decays from the time of tip detection according to a confidence decay coefficient. Let \( \lambda_i \in \mathbb{R}_+ \) denote a decay rate that reflects how quickly model certainty decreases over time for cue \( i \). Then, the desirability function may be modeled as:
\begin{equation}
\psi_i(t) = \exp\bigl(-\lambda_i t\bigr),
\label{eq:exp_decay}
\end{equation}
where \(t=0\) is the detection time \( n_i \), and the utility drops as time progresses. This exponential decay captures the intuition that observations are more valuable when acquired close to the time of detection, especially when forecasting uncertainty compounds rapidly. Such decay-aware task models have been applied in drone and constellation scheduling contexts~\cite{ramezani2023safe}.

In the maritime vessel scenario, for instance, dynamic cues are constructed using vessel motion predictions from AIS data, and utilities may be chosen to reflect either a predicted time of optimal visibility or a decaying confidence based on the growth of prediction error.

\begin{figure}[t]
  \centering
  \includegraphics[width=\columnwidth]{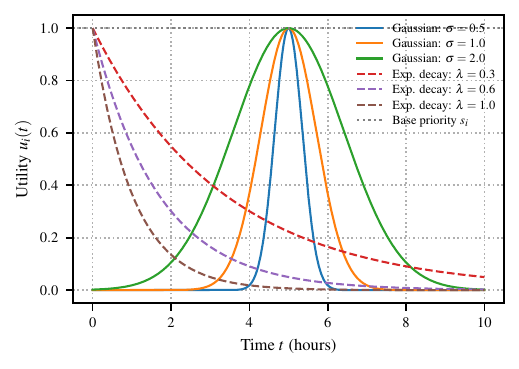}
  \caption{Illustration of utility functions \( u_i(t) = s_i \,\psi_i(t) \) with base priority \( s_i = 1.0 \). 
  Solid curves correspond to Gaussian-shaped utilities with peak time \( t_{\mathrm{peak}} = 5 \) hours and sharpness parameters \( \sigma \in \{0.5,\,1.0,\,2.0\} \). 
  Dashed curves represent exponential confidence decay functions with decay rates \( \lambda \in \{0.3,\,0.6,\,1.0\} \). 
  The base priority level \( s_i \) is shown as a horizontal reference line.}
  \label{fig:utility_curves}
\end{figure}

Figure~\ref{fig:utility_curves} shows utility curves with a fixed peak at \( t_{\text{peak}} = 5 \) and varying sharpness parameters \( \sigma_i \), as well as exponential decay curves based on confidence decay coefficients \( \lambda_i \). These illustrate two families of temporal desirability: symmetric peak-following and monotonic decay.

Feasibility constraints may also vary over time. For instance, daylight-only imaging may necessitate optical sensors, whereas nighttime acquisition requires SAR. Accordingly, each cue includes a time-dependent feasibility constraint set \( \mathcal{F}_i(t) \), defining the admissible sensor-task configurations as a function of time.

However, for optimization, we treat \( \mathcal{F}_i(t) \) as constant over time, denoted \( \mathcal{F}_i \), to ensure smoothness and tractability of the scheduling problem.

Each resulting cue is thus represented as:
\begin{equation}
c_i(t) = \bigl(\mathcal{P}_i(t),\, u_i(t),\, \mathcal{F}_i\bigr).
\end{equation}
which encapsulates the spatial target, the utility over time, and fixed operational feasibility constraints. These cues serve as inputs to the scheduling phase, where they are jointly optimized across satellites and observation windows.

\subsection{Task Scheduling Optimization}
\label{subsec:scheduling}

The task scheduling module selects acquisition times for a subset of generated cues by maximizing total utility while adhering to satellite-specific constraints. Positioned after cue generation, this stage determines which cues to image and when, using platform-specific observation windows and feasibility conditions. Unlike traditional slot-based scheduling, this formulation operates over disjoint continuous-time intervals, enabling fine-grained optimization via gradient methods.

\subsection*{Feasible Acquisition Windows}
\label{subsec:feasible_windows}

We consider a set of \( K \) satellites \( \{s_k\}_{k=1}^K \) with corresponding sensors \( \{\sigma_k\}_{k=1}^K \). The set of feasible acquisition times for cue \( i \) is denoted \( \mathcal{W}_i \subset \mathbb{R} \), defined as:
\begin{equation}
\mathcal{W}_i = \left\{ t \in \mathbb{R} \;\middle|\;
\begin{aligned}
& \exists\, k \in \{1, \dots, K\} \text{ such that } \\
& \mathcal{P}_i(t) \cap \mathcal{V}_{k}(t) \neq \emptyset \text{ and } \\
& \mathcal{F}_i \text{ is satisfied for sensor } \sigma_k
\end{aligned}
\right\},
\label{eq:wi_definition}
\end{equation}
where \( \mathcal{P}_i(t) \) is the predicted time-dependent footprint of cue \( i \), \( \mathcal{V}_k(t) \) is the visibility region of satellite \( s_k \) at time \( t \), and \( \mathcal{F}_i \) denotes time-invariant feasibility constraints such as sensor type, resolution, or angle requirements.

To compute \( \mathcal{W}_i \), we simulate satellite trajectories and intersect their viewing geometries with the predicted region \( \mathcal{P}_i(t) \) across time~\cite{wolfe2000three}. Because both satellite and target motions are continuous, visibility between two nearby times implies visibility in between, assuming the sampling is sufficiently dense. A sufficient sampling rate \( G \) (samples per unit time) can be estimated by either of the following two methods.

First, a theoretical approximation assumes that the satellite's velocity dominates the relative motion, i.e., \( v_{\text{sat}} \gg v_{\text{target}} \). In this case, the sampling rate must satisfy:
\begin{equation}
G > \frac{v_{\text{sat}}}{d_{\min}},
\end{equation}
where \( d_{\min} \) is the minimum footprint diameter, ensuring spatial continuity~\cite{wolfe2000three}.

However, this bound is typically too conservative in practice. Due to the finite imaging dwell time \( T_{\text{img}} \), a satellite cannot complete successive acquisitions more frequently than every \( T_{\text{img}} \) seconds. Thus, no new visibility opportunity can be physically realized between two samples separated by less than \( T_{\text{img}} \). As a result, it suffices to choose a practical sampling rate satisfying:
\begin{equation}
G > \frac{c}{T_{\text{img}}},
\end{equation}
for some small safety factor \( c > 1 \), such as \( c = 2 \) or \( c = 5 \). This ensures that the sampling resolution is physically meaningful and avoids oversampling time points that could not lead to additional feasible acquisitions due to the satellite’s operational latency.

Alternatively, an algorithmic test iteratively increases the sampling rate \( G \) until a fixed number of uniformly spaced intermediate points between each consecutive pair \( t_k, t_{k+1} \) are verified to satisfy the visibility condition across all intervals.

Each feasible interval of average duration \( T \) is uniformly discretized according to the sampling rate \( G \). For \( A \) access intervals per cue, the total number of sampling points per cue is \( A  G  T \). Additionally, each feasible window \( \mathcal{W}_i \) is sorted in time, and its \( A \) disjoint feasible intervals are explicitly stored as endpoint pairs, resulting in \( 2A \) boundary points per cue.

\noindent \textbf{Optimization Objective.}  
Let $s(i) \in \{1, \dots, K\}$ denote the satellite assigned to cue $i$, based on which satellite-sensor pair provides feasible access. The scheduling problem is formulated as:
\begin{equation}
\max_{\mathcal{S} \subseteq \{1, \dots, N\}} 
\left\{
\begin{array}{@{}c@{}}
\displaystyle \max_{\{t_i\}_{i \in \mathcal{S}}} \sum_{i \in \mathcal{S}} u_i(t_i) \\[1ex]
\text{s.t.} \\[0.5ex]
t_i \in \mathcal{W}_i \quad \forall i \in \mathcal{S} \\[0.5ex]
|t_i - t_j| \geq \Delta_{ij}(t_i, t_j) \quad \forall\ i,j \in \mathcal{S},\ s(i) = s(j)
\end{array}
\right\}.
\label{eq:objective}
\end{equation}

The nested formulation reflects that not all cues may be scheduled, and for those selected, acquisition times must both lie within their feasible windows and satisfy minimum time separation requirements~\cite{wolfe2000three, lemaitre2002agile}.

Here, $\Delta_{ij}(t_i, t_j)$ denotes the time-dependent buffer between imaging tasks, accounting for satellite repositioning and dwell time.  Since the angular separation $\gamma_{ij}(t_i, t_j)$ varies with both acquisition times according to the instantaneous spatial footprints of the targets, the buffer is estimated as:
\begin{equation}
\Delta_{ij}(t_i, t_j) = T_{\text{img}} + \frac{\gamma_{ij}(t_i, t_j)}{v_{\text{slew}}}
\label{eq:deltaij_timepair},
\end{equation}
where \( \gamma_{ij}(t_i, t_j) \) is the angular separation between targets \( i \) and \( j \) at times \( t_i \) and \( t_j \), respectively, 
\( v_{\text{slew}} \) is the satellite slew rate (e.g., degrees/sec), 
and \( T_{\text{img}} \) is the time required to acquire a single observation~\cite{zhang2014multi}. \( T_{\text{img}} \) may vary depending on the footprint size or scene type (e.g., larger targets may require longer dwell times), but for tractability, it is treated as a constant. Like the slew rate \( v_{\text{slew}} \), it is either provided by the satellite operator or chosen as a conservative upper bound when unavailable.

\noindent \textbf{Optimization Method.}  
To guide the initial selection of cues before scheduling, we employ an enhanced greedy pre-selection strategy that ranks each cue $i$ by:
\begin{equation}
\mathrm{rank}_i = \lambda\, \mathrm{avl}_i + (1 - \lambda) 
\max_{t \in \mathcal{W}_i} u_i(t),
\qquad \lambda \in [0, 1],
\label{eq:rank}
\end{equation}

where $\mathrm{avl}_i  \in [0, 1]$ quantifies the temporal uniqueness or sparseness of the cue. We define availability as
\begin{equation}
\mathrm{avl}_i = 1 - \frac{1}{N - 1} 
\sum_{j \neq i} \frac{\lvert \mathcal{W}_i \cap \mathcal{W}_j \rvert}{\lvert \mathcal{W}_i \rvert}.
\label{eq:availability}
\end{equation}
This represents the average normalized overlap between cue $i$ and all other cues. The parameter $\lambda \in [0, 1]$ balances utility with temporal availability.

We define the total penalty due to insufficient separation between tasks as:
\begin{equation}
\mathcal{P}_{\mathcal{S}} = 
\sum_{i < j} \kappa\bigl(\lvert t_i - t_j \rvert, \Delta_{ij}(t_i, t_j)\bigr)\,
\mathbbm{1}\bigl(s(i) = s(j)\bigr),
\label{eq:penalty}
\end{equation}
where \( \mathbbm{1}(\cdot) \) denotes the indicator function, and the soft penalty function \( \kappa(d, \Delta) \) is defined as:
\begin{equation}
\kappa(d, \Delta) = 
\max\Bigl\{ 0,\ \bigl[ 1 - (d/\Delta)^r \bigr]^{\beta} \Bigr\}.
\label{eq:kappa}
\end{equation}

This penalty is large when the time difference \( d = |t_i - t_j| \) is much smaller than the required separation \( \Delta_{ij} \), and decays smoothly to zero as \( d \to \Delta_{ij} \). The exponents \( r > 0 \) and \( \beta > 0 \) control the shape of the decay~\cite{nocedal2006numerical, dolgopolik2016smooth}.

\begin{figure}[t]
  \centering
  \includegraphics[width=\columnwidth]{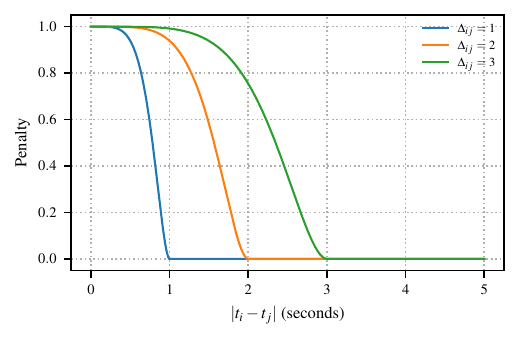}
  \caption{Soft penalty function \( \kappa(|t_i - t_j|, \Delta_{ij}) \) with parameters \( r = 5 \) and \( \beta = 2 \), shown for different buffer values \( \Delta_{ij} \in \{1,\,2,\,3\} \). 
Smaller \(\Delta_{ij}\) values correspond to sharper penalty decay with respect to the temporal separation \(|t_i - t_j|\). 
In this example, \(\Delta_{ij}\) is constant, corresponding to fixed spatial polygons for cues \( i \) and \( j \).}
  \label{fig:penalty_plot}
\end{figure}

An example of this soft penalty function, using the values \( r = 5 \) and \( \beta = 2 \), 
is shown in Figure~\ref{fig:penalty_plot}, demonstrating how it behaves across different values of \( \Delta_{ij} \). In this illustrative example, \( \Delta_{ij} \) is kept constant, corresponding to fixed spatial polygons for cues \( i \) and \( j \), whereas in the general formulation, \( \Delta_{ij}(t_i, t_j) \) may vary over time as the target footprints evolve.

Then the penalized loss function becomes:
\begin{equation}
\mathcal{L}(t) = -\, \sum_{i \in \mathcal{S}} u_i(t_i) + \rho\, \mathcal{P}_{\mathcal{S}}.
\label{eq:loss}
\end{equation}

The first term penalizes lower expected utility. The second term penalizes insufficient separation between tasks on the same satellite, encouraging temporal spacing to satisfy the buffer $\Delta_{ij}(t_i, t_j)$. The hyperparameter $\rho$ controls the penalty strength and is selected to be significantly larger than the maximum gradient scale of the utility terms~\cite{bertsimas1997introduction, boyd2004convex}. This ensures that penalty violations dominate the objective, thereby enforcing feasibility when a feasible configuration exists.

We begin with all generated cues and apply projected gradient descent to minimize the loss function over their acquisition times~\cite{nocedal2006numerical}. Gradient descent is used due to the continuous and differentiable nature of the utility and penalty terms, which allows efficient convergence to a locally optimal acquisition schedule. The projection step ensures that all updates remain within the feasible acquisition windows $\mathcal{W}_i$. If infeasible scheduling arises (i.e., penalty $\mathcal{P}_{\mathcal{S}} > 0$), we perform binary search over prefix sizes of the ranked cue list until feasibility is achieved~\cite{mcclellan1974art}. 

Afterward, we execute a final refinement phase. Each remaining unscheduled cue is re-evaluated in its original pre-order ranking and may be added back to the schedule if a feasible acquisition time can be found that does not conflict with any already scheduled cue. This postprocessing ensures that all cues, regardless of whether they were initially excluded, are reconsidered.

\noindent \textbf{Initialization and Refinement Time Selection.}  
In both the initial scheduling and the final refinement phases, an acquisition time $t_i$ must be selected for each cue by maximizing its utility function over the set of feasible times.

In the initialization phase, we first partition the feasible acquisition window $\mathcal{W}_i$ into continuous access intervals $\mathcal{I} \subset \mathcal{W}_i$. If no valid access interval $\mathcal{I}$ exists, the cue is excluded. 

In the refinement phase, each interval $\mathcal{I}$ is further divided into feasible sub-subintervals $\mathcal{J} \subset \mathcal{I}$ that additionally satisfy the temporal separation constraint $|t_i - t_j| \geq \Delta_{ij}(t_i,t_j)$ for all previously scheduled cues $j$ on the same satellite. If no such feasible $\mathcal{J}$ exists, the cue is excluded.

Within each interval or sub-subinterval, depending on the phase, we differentiate $u_i(t)$, locate critical points, and check whether they correspond to local maxima. If so, they are evaluated; otherwise, the maximum among the two boundaries is considered. The global maximum over all valid segments is selected.

\noindent \textbf{Convergence and Correctness.}  
We verify the convergence of the projected gradient descent (PGD) algorithm under the assumptions for nonconvex prox-regular constraint sets~\cite{barber2018gradient}. Specifically, we demonstrate that all necessary conditions for convergence to a stationary point are met in our setup. By Theorem 1 of~\cite{barber2018gradient}, this ensures that the iterations \( t^{(k)} \) produced by the PGD algorithm satisfy
\begin{equation}
\lim_{k \to \infty} \bigl\lVert \nabla \mathcal{L}\bigl(t^{(k)}\bigr) \bigr\rVert = 0.
\label{eq:cinvergence}
\end{equation}
and thus converge to a stationary point of the constrained problem.

The objective function consists of utility and penalty components. Assume that each utility function \( u_i(t) \) is differentiable and \( L_i \)-Lipschitz smooth. Define:
\begin{equation}
L = \max_i L_i, \qquad L' = \max\bigl(L, L_{\mathrm{pen}}\bigr).
\label{eq:lipschitz_constants}
\end{equation}

where \( L_{\mathrm{pen}}\) is the Lipschitz constant of the  soft penalty term. Therefore, the total objective \( \mathcal{L}(t) \) is globally \( L' \)-Lipschitz smooth~\cite{boyd2004convex, nocedal2006numerical}.

Assume further that each feasible set \( \mathcal{W}_i \) is a finite union of closed intervals:
\begin{equation}
\mathcal{W}_i = \bigcup_{a=1}^{A} \bigl[ a_{ia},\, b_{ia} \bigr].
\label{eq:wi_union}
\end{equation}

The corresponding projection operator is defined as:
\begin{equation}
\Pi_{\mathcal{W}_i}(t) = \arg \min_{\tau \in \mathcal{W}_i} \lvert t - \tau \rvert.
\label{eq:projection_operator}
\end{equation}

We now verify the five convergence conditions required for PGD:

\begin{enumerate}
\item \textbf{Smoothness:} The total objective \( \mathcal{L}(t) \) is differentiable and globally \( L' \)-Lipschitz smooth, as established by Equation~\eqref{eq:lipschitz_constants}. Therefore, the gradient \( \nabla \mathcal{L}(t) \) exists and is bounded~\cite{boyd2004convex}.

\item \textbf{Projection Operator:} Each projection \( \Pi_{\mathcal{W}_i} \) maps \( t \) to its nearest point in a finite union of intervals. It is piecewise linear and 1-Lipschitz (non-expansive), except at the endpoints of the interval (a measure-zero set). The full projection \( \Pi_{\mathcal{W}} = \Pi_{\mathcal{W}_1} \times \cdots \times \Pi_{\mathcal{W}_N} \) is separable and inherits this local Lipschitz continuity~\cite{rockafellar1998variational}.

\item \textbf{Prox-Regularity:} Each \( \mathcal{W}_i \) is a finite union of disjoint closed 1D intervals. Under mild separation (i.e., non-touching intervals), such sets are prox-regular~\cite{drusvyatskiy2018error}. Cartesian products of prox-regular sets remain prox-regular, so the full domain \( \mathcal{W} = \mathcal{W}_1 \times \cdots \times \mathcal{W}_N \) is prox-regular at any limit point.

\item \textbf{Step Size:} We enforce the condition \( \eta < \frac{1}{L'} \) explicitly in our implementation. This guarantees that each update step is appropriately scaled relative to the objective's curvature~\cite{nocedal2006numerical}.

\item \textbf{Neighborhood Stability:} Around any limit point \( t^* \), the projection operator \( \Pi_{\mathcal{W}} \) is stable and single-valued under sufficiently small steps, ensuring that PGD iterates remain within a consistent neighborhood~\cite{barber2018gradient}. While iterates may traverse different intervals of \( \mathcal{W}_i \) in the early steps, convergence occurs within a single connected component.

\end{enumerate}

\noindent \textbf{Complexity and Rate of Convergence.}  
We analyze the computational complexity of the key steps involved in the cue processing and scheduling pipeline, assuming \( N \) cues, each with \( A \) access intervals of average duration \( T \), and a sufficient temporal sampling rate \( G \).

The cost of computing the feasible acquisition window \( \mathcal{W}_i \) for each cue is dominated by intersecting satellite footprints with \( A G T \) sampled time points. Thus, the total cost for all \( N \) cues is:
\begin{equation}
\mathcal{O}(N A G T) = \mathcal{O}(G T).
\label{eq:feasible_acquisition_time_complexity}
\end{equation}

since \( N \) and \( A \) are assumed small relative to the temporal granularity \( G T \).

For each cue, evaluating the maximum utility value \( \max_{t \in \mathcal{W}_i} u_i(t) \) over its \( A \) intervals has cost \( \mathcal{O}(A) \), and thus across all cues:
\begin{equation}
\mathcal{O}(N A)
\label{eq:maximum_utility_time_complexity}
\end{equation}

Availability scores require computing normalized overlaps between the \( A \) intervals of each cue and those of all other \( N-1 \) cues. This yields a total cost:
\begin{equation}
\mathcal{O}(N^2 A^2)
\label{eq:availability_scores_time_complexity}
\end{equation}

Ranking and sorting all cues by their computed priority scores has complexity:
\begin{equation}
\mathcal{O}(N \log N)
\label{eq:sorting_time_complexity}
\end{equation}

The binary search stage applies projected gradient descent to prefixes of the cue list. Each evaluation involves: computing utility gradients in \( \mathcal{O}(N) \), penalty gradients in \( \mathcal{O}(N^2) \), and projections into \( \mathcal{W}_i \), each requiring \( \mathcal{O}(\log A) \) and totaling \( \mathcal{O}(N \log A) \)~\cite{boyd2004convex, nocedal2006numerical}.

Projected gradient descent converges sublinearly under fixed step size \( \eta \in (0, \frac{1}{L'}) \), requiring
\begin{equation}
\mathcal{O}\biggl(\frac{1}{\epsilon}\biggr) \ \text{iterations to reach}\ 
\lVert \nabla \mathcal{L}(t) \rVert \leq \epsilon
\label{eq:convergence_time_complexity}
\end{equation}
according to classical convergence bounds for nonconvex smooth optimization~\cite{barber2018gradient}. 

Hence, the total complexity of the binary search over all prefixes is:
\begin{equation}
\mathcal{O}\biggl( \frac{ \bigl( N^2 + N \log A \bigr) \log N }{ \epsilon } \biggr)
\label{eq:binary_search_total_time_complexity}
\end{equation}

During the final refinement phase, each unscheduled cue is checked for feasibility against the \( N \) already scheduled cues. For each scheduled cue, the potentially overlapping subinterval in the current cue is located in \( \mathcal{O}(\log A) \), and conflicting regions are removed. This produces up to \( \mathcal{O}(A + N) \) remaining intervals. Since each interval is scored in constant time, the total complexity across all cues is:
\begin{equation}
\mathcal{O}(N^2 \log A + N A + N^2)
\label{eq:refinment_time_complexity}
\end{equation}

\noindent \textbf{Optimization Algorithm.}
\begin{algorithm}[H]
\caption{Projected Gradient Scheduling with Pre-Selection}
\label{alg:gradient_scheduler}
\begin{algorithmic}[1]
\Require Cues $\{c_i\}_{i=1}^N$, step size $\eta$, penalty $\rho$, threshold $\epsilon$, balance weight $\lambda$, max gradient iterations $M$
\State $\mathcal{S} \gets \emptyset$
\For{each cue $i = 1$ to $N$}
  \State Compute feasible window $\mathcal{W}_i$
  \If{$\mathcal{W}_i = \emptyset$} \State \textbf{continue} \EndIf
  \State Sort all sampled time points in $\mathcal{W}_i$ and save the $2A$ interval endpoints
  \State Compute $\max_{t \in \mathcal{W}_i} u_i(t)$
  \State Compute $\text{avl}_i$
  \State Compute $r_i \gets \lambda\  \text{avl}_i + (1 - \lambda) \max u_i(t)$
  \State Add $c_i$ to $\mathcal{S}$
\EndFor
\State Sort cues in $\mathcal{S}$ in descending order of $r_i$
\State Initialize $k_{\text{min}} \gets 1$, $k_{\text{max}} \gets |\mathcal{S}|$, $k^* \gets 0$
\While{$k_{\text{min}} \leq k_{\text{max}}$}
  \State $k \gets \lfloor (k_{\text{min}} + k_{\text{max}})/2 \rfloor$
  \State $\mathcal{S}_k \gets$ top $k$ cues in the sorted list
  \For{each $i \in \mathcal{S}_k$}
    \State Initialize $t_i^{(0)} \gets \arg\max_{t \in \mathcal{W}_i} u_i(t)$
  \EndFor
  \For{$m = 0$ to $M$}
    \State Compute $\nabla \mathcal{L}(t^{(m)})$
    \State Update $\tilde{t}^{(m+1)} \gets t^{(m)} - \eta \cdot \nabla \mathcal{L}(t^{(m)})$
    \State Project: $t_i^{(m+1)} \gets \arg\min_{t \in \mathcal{W}_i} |t - \tilde{t}_i^{(m+1)}|$
    \If{$\|\nabla \mathcal{L}(t^{(m+1)})\| < \epsilon$}
      \State \textbf{break}
    \EndIf
  \EndFor
  \If{$\mathcal{P}_{\mathcal{S}_k} = 0$}
    \State $k^* \gets k$
    \State $k_{\text{min}} \gets k + 1$
  \Else
    \State $k_{\text{max}} \gets k - 1$
  \EndIf
\EndWhile
\State $\mathcal{S}_{k^*} \gets$ top $k^*$ cues in the sorted list
\For{each unscheduled $c_i \notin \mathcal{S}_{k^*}$ in pre-order}
  \State $t_i^* \gets \arg\max_{t \in \mathcal{W}_i \text{ s.t. } |t_i - t_j| \geq \Delta_{ij},\ \forall j \in \mathcal{S}_{k^*}} u_i(t)$
  \If{feasible $t_i^*$ exists} \State Add $c_i$ to $\mathcal{S}_{k^*}$
  \EndIf
\EndFor
\State \Return $\{t_i\}_{i \in \mathcal{S}_{k^*}}$
\label{alg:optimization_alg},
\end{algorithmic}
\end{algorithm}

\noindent \textbf{Optimization Metric.}  
The final schedule is evaluated by the total utility:
\begin{equation}
U_{\mathrm{tot}} = \sum_{i \in \mathcal{S}_{k^*}} u_i(t_i)
\label{eq:optimization_metric}
\end{equation}
which reflects the aggregate expected value of all scheduled observations in the selected subset.

\subsection{Imaging Execution}
\label{subsec:execution}

Following the optimization phase, imaging tasks are executed only for cues that were successfully assigned a feasible acquisition time and matched to a valid satellite–sensor platform. Each selected cue is represented as

\begin{equation}
c_i(t_i) = \bigl( \mathcal{P}_i(t_i),\, u_i(t_i),\, \mathcal{F}_i \bigr)
\label{eq:scheduled_cue}
\end{equation}

where \( t_i \in \mathcal{W}_i \) and all constraints in \( \mathcal{F}_i \) are satisfied.

For each \( c_i(t_i) \) in the selected group, a tasking request is generated. This includes the spatial footprint, sensor mode, acquisition time \( t_i \), and platform assignment. Tasking requests are submitted to the relevant satellite operator, either via commercial APIs, simulation environments, or direct-access interfaces.

The framework monitors execution feedback for each request, including acknowledgment of receipt, confirmation of successful acquisition, or notification of failure. All metadata is logged to support subsequent analysis and error handling in downstream modules.

\subsection{Image Analysis and Enrichment}
\label{subsec:image_analysis}

Each acquired image \( \boldsymbol{{I_{n_i}}} \) is processed through an enrichment operator \( \mathcal{E} \) to extract semantic and contextual information. This yields a structured representation:
\begin{equation}
\boldsymbol{d_{n_i}} = \mathcal{E}\bigl(\boldsymbol{I_{n_i}}\bigr)
\label{eq:enriched_imaged}
\end{equation}
where in this case, \( \boldsymbol{{d_{n_i}}} \) may include outputs from detectors, vision–language models, or other pretrained AI models~\cite{radford2021learning, li2022blip, liu2024remoteclip, xiao2025foundation}.

To evaluate semantic changes over time, the enriched result is compared with its historical context \( \boldsymbol{{H_{n_i}}} \) using a comparison operator \( \mathcal{M} \):
\begin{equation}
\boldsymbol{\delta_{n_i}} = \mathcal{M}\bigl(\boldsymbol{d_{n_i}}, \boldsymbol{H_{n_i}}\bigr) =
\bigl[ \delta_{n_i}^{(1)},\, \delta_{n_i}^{(2)},\, \dots,\, \delta_{n_i}^{(K)} \bigr]
\label{eq:semantic_changes}
\end{equation}

with each component \( \delta_{n_i}^{(k)} \) capturing a distinct deviation or shift~\cite{li2021deep, bai2023deep}.

These are aggregated into a scalar relevance score:
\begin{equation}
r_{n_i} = f\bigl(\boldsymbol{\delta_{n_i}}\bigr) = \sum_{k=1}^{K_i} w_k\, \delta_{n_i}^{(k)}, 
\qquad \sum_{k=1}^{K_i} w_k = 1
\label{eq:relevance_score}
\end{equation}

The resulting \( r_{n_i} \in [0,1] \) is passed to the reporting module for decision-making and feedback generation~\cite{cao2025open}.

\subsection{Report Generation and Feedback}
\label{subsec:report_feedback}

This module generates structured summaries for each acquired image and manages feedback into the tip extraction phase~\cite{logar2020pulsesatellite}. For every discrete time step \( n \), the system compiles the detected objects, semantic outputs, comparative insights based on \( \boldsymbol{{d_{n_i}}} \), and the relevance score \( r_{n_i} \in [0,1] \) from the previous analysis stage.

These elements are combined into a visual and semantic report~\cite{wu2024trtr, zhang2024rs5m}, which supports both automated pipelines and human interpretation. Such reporting pipelines are critical in operational systems to enhance decision support and update tasking logic with minimal latency.

To determine whether a new tip should be generated from the image at time \( n \), the system applies a binary decision rule over the relevance score:
\begin{equation}
f_{\tau_i}(r_{n_i}) =
\begin{cases}
1, & \text{if } r_{n_i} > \theta_i, \\[6pt]
0, & \text{otherwise.}
\end{cases}
\label{eq:decision_rule_tip}
\end{equation}

If triggered, the image’s metadata is used to define a new tip, which is passed back to the tip extraction module. This mechanism complements externally sourced tips and enables adaptive downstream tasking based on recent observations, enabling closed-loop learning in spaceborne monitoring systems.

\section{Experimental Setup}
\label{sec:experimental_setup}

To evaluate the proposed Tip-and-Cue framework in a realistic maritime context, we conduct experiments using the TrAISformer model~\cite{traisformer2024} for trajectory forecasting and tip generation. The study area is located off the U.S. East Coast, bounded by latitude $[39.8^\circ, 41.0^\circ]$ and longitude $[-74.4^\circ, -72.5^\circ]$. Vessel AIS data was collected from the publicly available NOAA AIS archive~\cite{noaaais2024}.

\subsection{Training, Validation, and Test Periods for the TrAISformer}
\label{subsec:TrAISformer_model}

The TrAISformer model is trained on AIS data within the specified polygon from January~1,~2024 to March~4,~2024. 
Validation is conducted from March~5 to March~10, and testing spans March~11 to March~31. 
Table~\ref{tab:mse_results} summarizes the mean squared error (MSE) in kilometers across prediction horizons ranging from 1 to 6 hours. 
The results show a clear trend of increasing error as the prediction horizon extends, which is expected in trajectory forecasting tasks. 
Overall, the performance remains consistent across the training, validation, and test sets, indicating stable generalization and reliable long-range prediction behavior.

\begin{table}[t]
\centering
\caption{Mean squared error (MSE) in kilometers after 1--6 hours for training, validation, and test sets.}
\label{tab:mse_results}
\begin{tabular}{ccccccc}
\hline
\textbf{Set} & \textbf{1h} & \textbf{2h} & \textbf{3h} & \textbf{4h} & \textbf{5h} & \textbf{6h} \\
\hline
Train        & 1.080 & 2.381 & 3.908 & 5.828 & 7.262 & 7.877 \\
Validation   & 1.366 & 2.129 & 3.784 & 5.804 & 7.805 & 8.657 \\
Test         & 1.480 & 2.816 & 4.831 & 6.594 & 6.776 & 7.162 \\
\hline
\end{tabular}
\end{table}

\subsection{Tips Gathering and Cues for Anomalous Vessel Trajectories}
\label{subsec:dynamic_tips_and_cues}

We focus on March 30, 2024. Dynamic tips are generated by comparing each AIS timestamp up to 18:30 UTC with the corresponding prediction based on AIS data up to one hour earlier. A tip is generated if the location error exceeds $\theta_{\text{AIS}} = 3.0\,\text{km}$. This process yields four dynamic tips on that day. Each tip \( \tau_i \) is scored using the schema defined in \eqref{eq:scoring_AIS}, with parameters $\Delta_{\text{lead},i} = 3.0\,\text{hrs}$, and $\alpha = 0.5$.

Each dynamic cue \( c_i \) corresponds to a square of $200 \times 200$ meters centered on the location of the vessel at each prediction time. The utility function for these cues is modeled by the exponential decay defined in~\eqref{eq:exp_decay} with $\lambda_i = 0.2$, and utility values below $10^{-3}$ are treated as zero.

\subsection{Synthetic Static Tips and Cues}
\label{subsec:static_tips_and_cues}

To represent additional areas of interest, we generate 100 synthetic static tips and cues. Their centers are drawn uniformly at random within the defined polygon. Each static cue is assigned a rectangular footprint with side lengths uniformly sampled from $[200, 800]$ meters, and a priority score \( s_i \) uniformly sampled from $[0.05, 0.25]$. The temporal desirability for each static cue follows the Gaussian utility function \(\psi_i(t)\) defined in Equation~\eqref{eq:gaussian_psi}, with $t_{\text{peak}}$ uniformly sampled between 18:30 and 23:59 UTC and $\sigma_i$ uniformly drawn from $[0.5\,\text{hr}, 2\,\text{hr}]$. As before, utility values below $10^{-3}$ are treated as zero.

\subsection{Cues Distribution and Constraints}
\label{subsec:cues_scatter_and_constraints}

\begin{figure}[t]
  \centering
  \includegraphics[width=\linewidth]{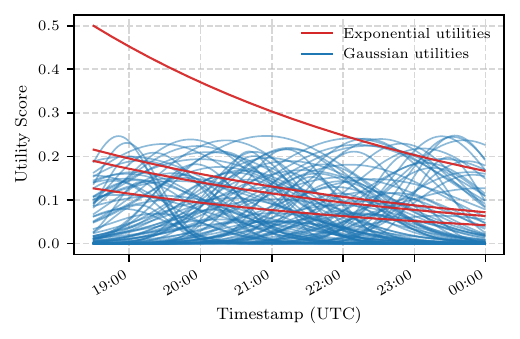}
  \caption{Utility functions \( \psi_i(t) \) for all cues on March 30, 2024, within the scheduling window from 18:30 to 23:59~UTC. Dynamic cues follow exponential decay, while static cues follow Gaussian utility models.}
  \label{fig:utilities_plot}
\end{figure}

All cue utility functions over the scheduling horizon are shown in Fig.~\ref{fig:utilities_plot}. Each curve represents the temporal desirability profile of a single cue. Dynamic cues exhibit exponential decay patterns, starting from their respective tip detection times, reflecting a decrease in model confidence over time. Static cues follow Gaussian-shaped profiles with varying peak times and sharpness parameters, corresponding to diverse temporal priorities randomly assigned within the scheduling window.

\begin{figure}[t]
  \centering
  \includegraphics[width=0.95\linewidth]{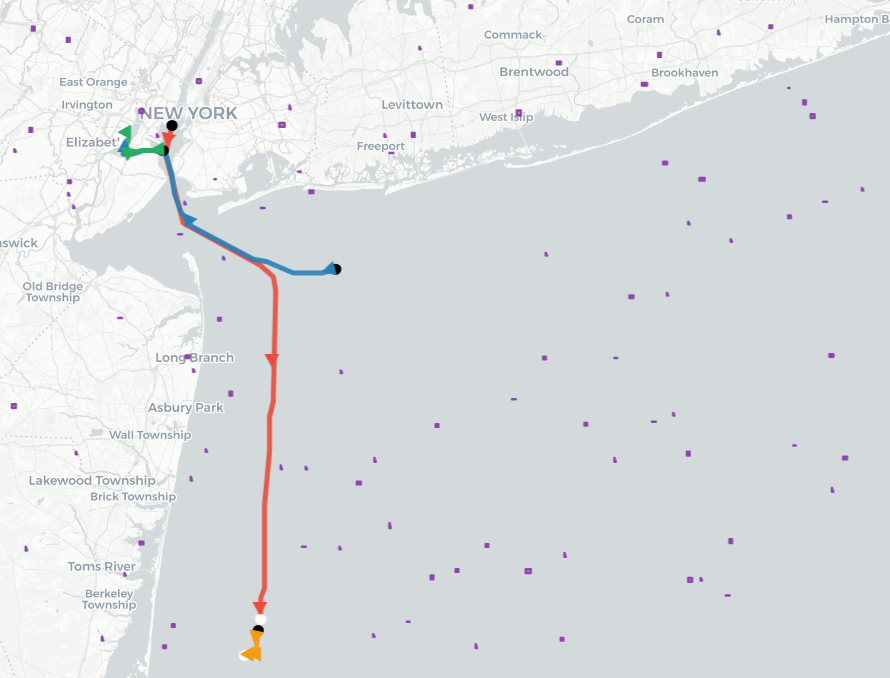}
  \caption{Spatial footprints of dynamic and static cues in the operational region of the U.S. East Coast. The 4 directed arrows represent the 4 dynamic cues, while the 100 purple rectangles represent the 100 static cues.}
  \label{fig:cue_map}
\end{figure}

The spatial distribution of all generated cues is shown in Fig.~\ref{fig:cue_map}. Each rectangle corresponds to the geographic footprint of a dynamic or static cue within the defined operational region. This map illustrates the diverse spatial characteristics of the cue pool used for scheduling.

All cues are assigned the following feasibility constraints for the scheduling process:
\begin{itemize}
    \item \texttt{Sensor Type} = EO
    \item \texttt{Max Cloud Cover} = \(100\,\%\)
    \item \texttt{Max Allowed GSD} = 100.0 (cm/px)
    \item \texttt{Max off Nadir Angle} = \(90.0^{\circ}\)

\end{itemize}

\subsection{Satellite Configuration}
\label{subsec:satellite_configuration}

We simulate three EO satellites: SKYSAT-C11, SKYSAT-C15~\cite{saunier2022skysat}, and JILIN-1-GF03D50~\cite{yu2020board, jianing2022research}. These satellites represent agile commercial platforms with submeter resolution and frequent revisit rates. Feasible acquisition windows are derived based on orbital parameters and the feasibility model in Section~\ref{subsec:feasible_windows}. 

Two-line element (TLE) sets for each satellite, corresponding to the relevant dates in March~2024, were obtained from the CelesTrak archive~\cite{kelso_celestrak}. Satellite trajectories were then propagated to the desired time intervals using the SGP4 orbital propagation model~\cite{vallado2006revisiting}, ensuring accurate visibility calculations for the operational region.

\subsection{Scheduling and Evaluation}
\label{subsec:scheduling_and_evaluation}

Scheduling is carried out using the procedure described in Algorithm~\ref{alg:gradient_scheduler}. The input consists of 104 cues and the orbital parameters for the three satellites. 

The quality of the resulting schedule is evaluated based on three aspects: the number of cues successfully scheduled, the total accumulated utility, and whether the resulting plan satisfies all visibility and temporal separation constraints. The total accrued utility is measured using the optimization metric defined in Equation~\eqref{eq:optimization_metric}, which quantifies the expected value of the final scheduled subset.

\section{Results}
\label{sec:results}

We evaluate the proposed tip-and-cue framework by applying the complete scheduling algorithm described in Section~\ref{subsec:scheduling} to the dynamic and static cues generated in Section~\ref{sec:experimental_setup}. The evaluation encompasses both quantitative performance metrics and a qualitative example that demonstrates the system's end-to-end capabilities.

\subsection{Scheduling Algorithm Summary}

The scheduling process follows Algorithm~\ref{alg:gradient_scheduler}, consisting of four main phases:  
(1) computation of feasible acquisition windows using the algorithmic sampling method described in Section~\ref{subsec:feasible_windows};  
(2) scoring and sorting of all cues according to their availability and maximum utility, as defined in Equation~\eqref{eq:rank};  
(3) selection of an optimal subset of cues via binary search, combined with projected gradient descent (PGD) over acquisition times to minimize the penalized loss defined in Equation~\eqref{eq:loss};  
and (4) a final refinement phase, which reconsiders previously unscheduled cues to add them if feasible.

The optimization uses the following hyperparameters and physical parameters: step size $\eta = 0.01$, penalty weight $\rho = 100.0$, convergence threshold $\epsilon = 0.001$, maximum number of iterations $M = 500$, and imaging dwell time $T_{\text{img}} = 1.0$~s. These values ensure stable PGD convergence and sufficiently fine temporal sampling for the considered satellite configuration.

\subsection{Feasible Acquisition Windows}

For the operational region off the U.S. East Coast, each of the three satellites has a single feasible window during the relevant time period (all times UTC):

\begin{itemize}
    \item \textbf{SKYSAT-C11:} 19:13–19:14 (80~s) 
    \item \textbf{SKYSAT-C15:} 18:56–18:58 (81~s) 
    \item \textbf{JILIN-1-GF03D50:} 20:05–20:06 (87~s)
\end{itemize}

Because these windows are temporally concentrated and the operational polygon is spatially small, the temporal overlap between satellite passes is limited. Out of the 104 cues generated, only 85 have a nonempty intersection between their feasible acquisition window and a positive-utility interval. This establishes an upper bound of 85 cues that can be scheduled in this experiment.

\subsection{Effect of $\lambda$ on Cue Ranking}

We assess the influence of the ranking weight $\lambda$ in Equation~\eqref{eq:rank}, which balances temporal availability against maximum utility. Choosing $\lambda = 0$ ranks cues solely by their utility, whereas $\lambda = 1$ ranks them exclusively by their temporal uniqueness. Intermediate values balance these two priorities.

\begin{table}[t]
\centering
\caption{Scheduling results for different $\lambda$ values, showing contributions of the binary search and refinement phases.}
\label{tab:lambda_results}
\begin{tabular}{ccccc}
\hline
$\lambda$ & Binary Search & Refinement & Total Scheduled & $U_{\text{total}}$ \\
\hline
0    & 67 & 14 & 81 & 8.9826 \\
0.25 & 65 & 17 & 82 & 8.9772 \\
0.5  & 65 & 17 & 82 & 8.9726 \\
0.75 & 67 & 16 & 83 & 8.9698 \\
1    & 66 & 15 & 81 & 8.9591 \\
\hline
\end{tabular}
\end{table}

Table~\ref{tab:lambda_results} summarizes the number of cues scheduled during the binary search and refinement phases, as well as the total number of scheduled cues and the overall utility defined in Equation~\eqref{eq:optimization_metric} for different $\lambda$ values. Moderate $\lambda$ values ($0.25 \leq \lambda \leq 0.75$) yield slightly higher total scheduled cues, with total utility remaining stable. The refinement phase consistently contributes between 14 and 17 additional scheduled cues, in addition to the binary search result.

\subsection{Loss Evolution During PGD}

\begin{figure}[t]
  \centering
  \includegraphics[width=\linewidth]{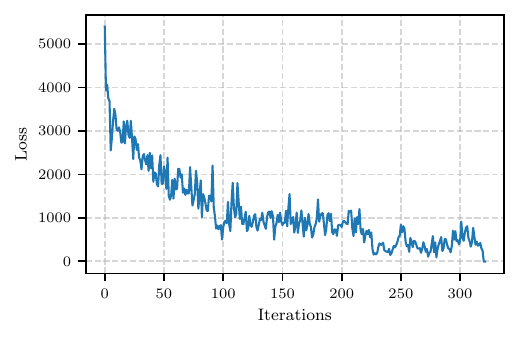}
  \caption{Loss value as a function of the iteration number of the Projected Gradient Descent for the optimal prefix size (65) during binary search, using \( \lambda = 0.25 \).}
  \label{fig:loss_iterations}
\end{figure}

Figure~\ref{fig:loss_iterations} illustrates the convergence behavior of the PGD algorithm during the binary search stage for $\lambda = 0.25$. The loss function decreases steadily and converges well within the allowed iteration budget, indicating that the chosen hyperparameters provide a good balance between convergence speed and stability.

\subsection{Qualitative Example}
\label{subsec:qualitative_example}

To illustrate the feasibility and semantic potential of the proposed framework, we present a qualitative example centered on the vessel \textit{HMM~GARNET} (IMO:~9944455), a container ship sailing under the flag of Liberia. This case study focuses on verifying whether the framework can generate a cue from predicted metadata, match it to a real Earth Observation (EO) satellite pass, and perform downstream enrichment using vision–language analysis.

A dynamic tip was generated by the TrAISformer model using AIS trajectory predictions, as described in Section~\ref{sec:experimental_setup}. A corresponding cue was formulated by defining a 200$\times$200~m square around the vessel's predicted location at each timestamp. The cue's spatial–temporal properties were checked for compatibility with real satellite overpasses using the feasibility logic outlined in Section~\ref{subsec:feasible_windows}.

\begin{figure*}[t]
  \centering
  \begin{subfigure}[t]{0.49\textwidth}
    \centering
    \includegraphics[width=\linewidth,height=0.34\textheight,keepaspectratio]{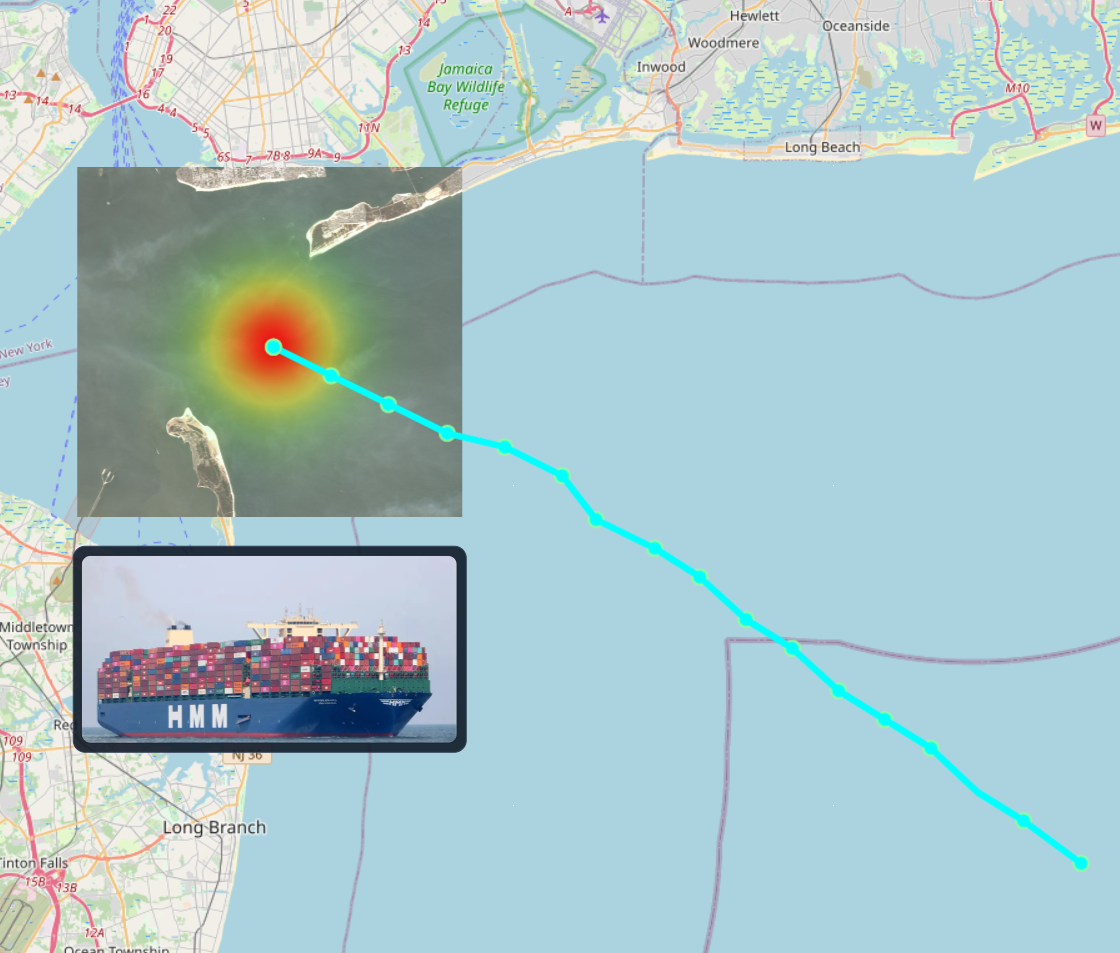}
    \caption{Predicted trajectory points of \textit{HMM~GARNET} up to the imaging time, overlaid on a map with the Sentinel-2 image, model heatmap, and ground photograph of the vessel.}
  \end{subfigure}
  \hfill
  \begin{subfigure}[t]{0.49\textwidth}
    \centering
    \includegraphics[width=\linewidth,height=0.34\textheight,keepaspectratio,clip,trim=5 5 5 5]{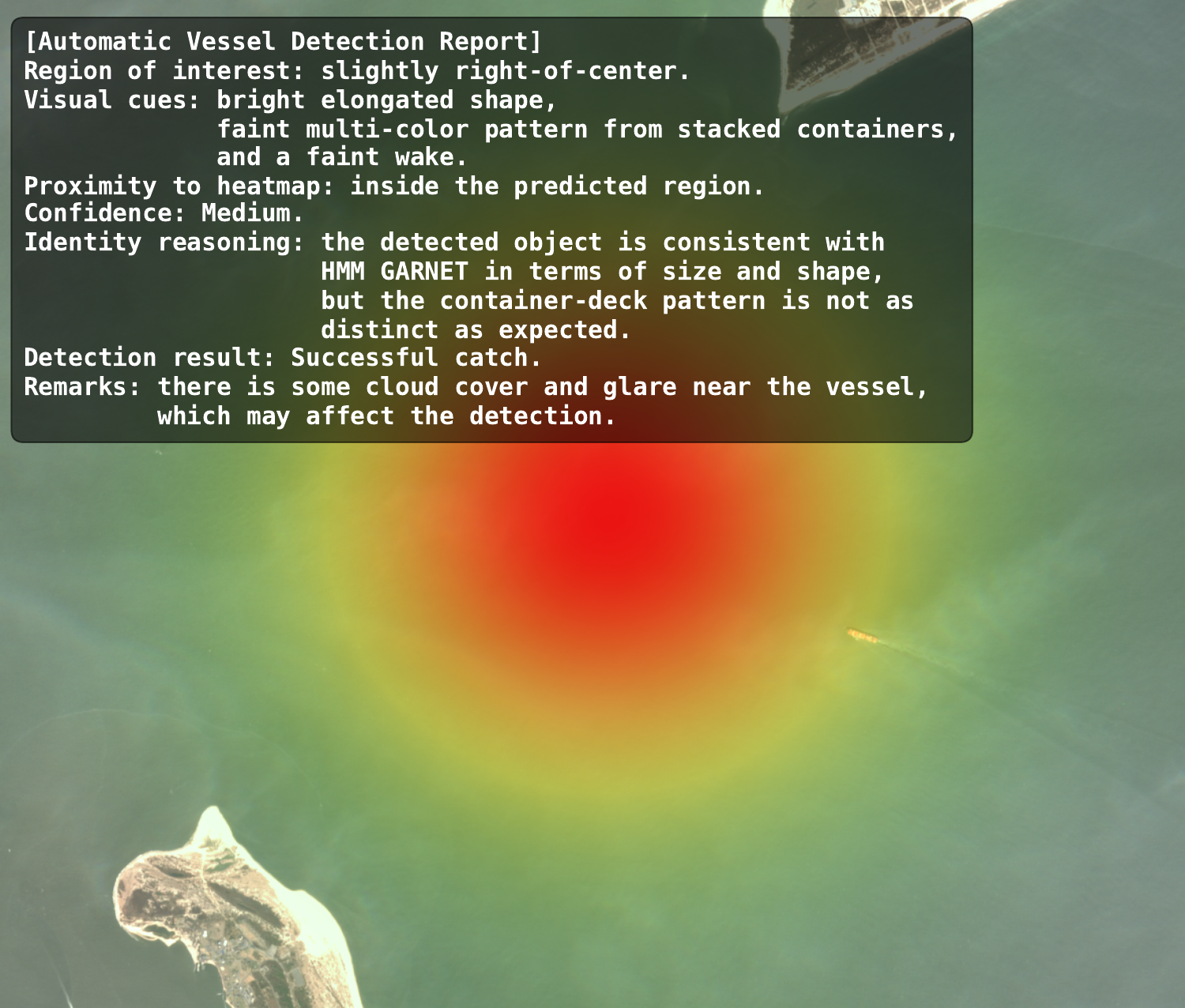}
    \caption{Crop of the Sentinel-2 image with the heatmap overlay and the corresponding VLM-generated visual report.}
  \end{subfigure}

  \caption{Qualitative example of the proposed framework applied to the vessel \textit{HMM~GARNET} (IMO:~9944455). 
  The model predictions and satellite imagery demonstrate end-to-end integration from AIS-based trajectory forecasting to feasible imaging and semantic interpretation.}
  \label{fig:qualitative_example}
\end{figure*}

A matching Sentinel-2 satellite pass was identified for the relevant time interval, indicating that this cue could have been feasibly imaged under the system’s assumptions. The left panel in Fig.~\ref{fig:qualitative_example} shows the predicted trajectory points of \textit{HMM~GARNET} up to the imaging time, overlaid on a map together with the corresponding Sentinel-2 image, the TrAISformer heatmap, and a reference ground-truth photograph of the vessel. The right panel provides a zoomed-in view of the same Sentinel-2 scene, overlaid with a heatmap and a caption generated by the pre–trained vision–language model (VLM).

To further demonstrate the framework’s potential for semantic enrichment, the NVIDIA ViLA model~\cite{lin2024vila} was applied to the Sentinel-2 image to produce a natural-language description of the detected vessel and surrounding scene. The resulting visual and textual outputs jointly illustrate the framework’s ability to integrate predictive tasking, feasible imaging, and high-level semantic interpretation.

\section{Discussion}
\label{sec:discussion}

The proposed Tip-and-Cue framework presents a scalable and fully automated solution for end-to-end satellite tasking, cue prioritization, schedule optimization, and image analysis. Our formulation addresses key challenges inherent in operational Earth Observation (EO) systems, including real-time tip responsiveness, continuous-time optimization under nonlinear utility functions, sensor-task feasibility constraints, and feedback-driven enrichment. Unlike traditional pipelines, which treat detection, scheduling, and analysis as disjoint modules, our unified architecture enables tight coupling across stages, where each component informs and enhances the performance of the downstream stages.

By formulating task scheduling as a constrained continuous optimization problem, our approach departs from classical discrete or rule-based methods. It demonstrates that projected gradient techniques can be used to optimize over nonconvex, time-dependent windows efficiently. This opens the door to handling richer utility functions and tighter geometric constraints than typically supported in existing systems. Prior work in continuous satellite scheduling focused on simplified scenarios such as single-satellite imaging or assumed fixed targets, whereas our method generalizes to multi-satellite, multi-constraint environments with time-varying polygonal footprints and physical buffer requirements.

Experimental results confirm the benefits of our joint optimization and refinement phases, showing improved schedule feasibility, cue utility, and spatial coverage. The soft-penalty formulation offers a flexible means to enforce satellite constraints without prematurely excluding high-utility cues. Furthermore, the integration of vision-language models and semantic feedback mechanisms supports closed-loop reasoning, enabling downstream imagery to adaptively inform future tasking decisions, a capability not addressed in most operational frameworks.

Future research should explore the principled design of utility functions that better capture domain-specific objectives such as change detection sensitivity, revisit prioritization, or multi-modal correlation. 
Additionally, while we treat feasibility constraints as time-invariant within the optimization loop, many real-world constraints (e.g., illumination conditions, cloud forecasts, or angle thresholds) are inherently time-dependent and could be incorporated into the objective via differentiable surrogate models. 
Another promising direction is the support for real-time cue updates or insertions during the optimization process, allowing the scheduler to adapt dynamically to streaming tips or high-priority events with low latency. 
More broadly, the proposed formulation can serve as a unified framework for future studies to evaluate the contribution and interaction of each component, such as cue generation, feasibility estimation, or scheduling optimization, while assessing overall end-to-end performance. 
It is also worth noting that this work does not explicitly address imaging geometries such as strip, spot, or mosaic acquisition modes, which represent a major source of complexity in other scheduling algorithms and should be systematically examined within this framework in future extensions.

\section{Conclusions}
\label{sec:conclusion}

We have presented a fully automated tip-and-cue framework for end-to-end satellite tasking, scheduling, and image analysis. 
The framework integrates all major phases, including tip extraction, cue generation, feasible window estimation, continuous-time scheduling optimization, and AI-based semantic enrichment, within a unified mathematical and operational formulation.
This holistic design enables seamless information flow across modules, from early anomaly detection to final semantic interpretation, establishing a consistent bridge between predictive metadata, satellite feasibility, and high-level image understanding. 
Overall, these components form a scalable foundation for autonomous satellite operations across multi-sensor constellations.

The modular pipeline supports both reactive and proactive imaging workflows, where each acquired observation directly informs subsequent acquisition decisions. 
By coupling prediction, optimization, and analysis in a single closed-loop process, the framework provides a foundation for scalable, adaptive, and intelligence-driven Earth observation systems.

This work opens new directions for large-scale and near-real-time satellite mission planning. The proposed scheduling formulation is generic and extensible, allowing the framework to be adapted to diverse sensing tasks across various domains. Future efforts should explore advanced utility modeling, explicitly time-varying constraints, and dynamic cue insertion to fully support continuous, adaptive decision-making in operational satellite systems.

\section*{Acknowledgements}
\label{sec:acknowlegements}
We express our sincere gratitude to Avraham Cohen and Peleg Levy for their valuable consultation, insightful discussions, and continuous support throughout this research.

\bibliographystyle{IEEEtran}
\bibliography{references}

\end{document}